\documentclass[letterpaper, 10 pt, conference]{ieeeconf}  
\usepackage[backend=biber, style=numeric, sorting=none]{biblatex}
\usepackage{graphicx}
\usepackage{subcaption}
\usepackage{color,soul}
\usepackage{amsmath}
\usepackage{algorithm}
\usepackage{algorithmicx}
\usepackage{setspace}
\usepackage{cleveref}
\usepackage{tikz}

\newcommand\copyrighttext{%
  \footnotesize \textcopyright~2025 IEEE. Personal use of this material is permitted. Permission from IEEE must be obtained for all other uses, in any current or future media,
including reprinting/republishing this material for advertising or promotional purposes, creating new collective works, for resale or redistribution to servers
or lists, or reuse of any copyrighted component of this work in other works.}

\newcommand\copyrightnotice{%
    \begin{tikzpicture}[remember picture,overlay]
    \node[anchor=south,yshift=10pt] at (current page.south) {\fbox{\parbox{\dimexpr\textwidth-\fboxsep-\fboxrule\relax}{\copyrighttext}}};
    \end{tikzpicture}%
}

\crefname{equation}{Eq.}{Eqs.}
\crefname{figure}{Fig.}{Figs.}
\crefname{table}{Table}{Tables}

\AtEveryBibitem{\small
  \setlength{\itemsep}{0pt}
  \setlength{\parskip}{0pt}
  \setlength{\bibitemsep}{0pt plus 0.3ex}
  \setlength{\bibitemsep}{0pt}
  \setlength{\bibhang}{0pt}
  \setlength{\bibnamesep}{0pt}
  \setlength{\biblabelsep}{3pt}
}

\IEEEoverridecommandlockouts                              
\title{\LARGE \bf
From Code to Road: A Vehicle-in-the-Loop and Digital Twin-Based Framework for Central Car Server Testing in Autonomous Driving}

\author{Chengdong Wu$^{1}{*}$, Sven Kirchner$^{1}{*}$, Nils Purschke$^{1}{*}$, Axel Torschmied$^{2}$, Norbert Kroth$^{2}$, Yinglei Song$^{1}$, \\André Schamschurko$^{1}$, Erik Leo Haß$^{1}$, Kuo-Yi Chao$^{1}$, Yi Zhang$^{1}$, Nenad Petrovic$^{1}$, Alois C. Knoll$^{1}$
\thanks{\hspace*{-1em}$^{1}$ Technical University of Munich, Garching, Bayern, Germany}%
\thanks{\hspace*{-1em}$^{2}$ CARIAD SE, Germany}%
\thanks{\hspace*{-1em}$^{*}$ Equal contribution}%
\thanks{\hrule}
\thanks{\hspace*{-1em} A preliminary version of this work appeared in the Proceedings of the Driving the Future Symposium (DTF), 2025. The present paper substantially extends that work by presenting a complete Vehicle-in-the-Loop (ViL) test bench with real sensor integration, enhanced autonomous driving algorithm implementation, and comprehensive quantitative validation of the central car server.}
\thanks{\hspace*{-1em} This research was funded by the Federal Ministry of Research, Technology and Space (BMFTR) as part of the CeCaS project, FKZ: 16ME0800K.}
\thanks{\hspace*{-1em} The publication was written within the Shift2SDV project (GA number 101194245) which is supported by the Chips Joint Undertaking and its members, including top-funding by the national authorities of Austria, Denmark, Germany, Greece, Finland, Italy, Netherlands, Poland, Portugal, Spain, Turkey. Co-funded by the European Union. Views and opinions expressed are however those of the author(s) only and do not necessarily reflect those of the European Union or the Chips Joint Undertaking. Neither the European Union nor the granting authorities can be held responsible for them.}
}

\begin{document}
\makeatletter
\let\@oldmaketitle\@maketitle
\renewcommand{\@maketitle}{\@oldmaketitle
}
\makeatother
\maketitle
\copyrightnotice
\thispagestyle{empty}
\pagestyle{empty}

\begin{abstract}
Simulation is one of the most essential parts in the development stage of automotive software. However, purely virtual simulations often struggle to accurately capture all real-world factors due to limitations in modeling. To address this challenge, this work presents a test framework for automotive software on the centralized E/E architecture, which is a central car server in our case, based on Vehicle-in-the-Loop (ViL) and digital twin technology. The framework couples a physical test vehicle on a dynamometer test bench with its synchronized virtual counterpart in a simulation environment. Our approach provides a safe, reproducible, realistic, and cost-effective platform for validating autonomous driving algorithms with a centralized architecture. This test method eliminates the need to test individual physical ECUs and their communication protocols separately. In contrast to traditional ViL methods, the proposed framework runs the full autonomous driving software directly on the vehicle hardware after the simulation process, eliminating flashing and intermediate layers while enabling seamless virtual-physical integration and accurately reflecting centralized E/E behavior. In addition, incorporating mixed testing in both simulated and physical environments reduces the need for full hardware integration during the early stages of automotive development. Experimental case studies demonstrate the effectiveness of the framework in different test scenarios. These findings highlight the potential to reduce development and integration efforts for testing autonomous driving pipelines in the future.

\end{abstract}


\addtocounter{figure}{-1} 

\section{INTRODUCTION}
Autonomous driving has emerged as a prominent focus in the automotive industry. Complex distributed E/E architectures in the vehicle cannot meet the requirements of multiple automotive system properties, such as computing-intensive algorithms \cite{daimler}. Additionally, sensors used for autonomous driving functions and communication channels for V2X deployments enhance the sophistication of the E/E architecture. In most cases, vehicular functions are mapped to the respective ECUs, resulting in more than 100 ECUs with different functions \cite{state}. Furthermore, the standard vehicle development process requires regular updates to vehicle functions, and introducing new functions can result in the addition of new ECUs within a decentralized architecture. This approach leads to significant cost increases due to bulky wiring harnesses, as well as higher software complexity and a large number of software variants \cite{case}.

Therefore, a centralized structure is preferred for automated-driven vehicles to facilitate understanding and implementation. Enhanced computational power and hardware provisions for functional safety and security are necessary for a centralized structure. Other increasing requirements include improved communication networks, with higher bandwidth, real-time and traffic partitioning capabilities, fault-tolerance mechanisms, advanced gateways, and enhanced security measures \cite{case}. The centralized structure does not require adding ECUs for new applications. Instead, it adopts a service-oriented approach, which is a base enabler to the Software-Defined Vehicle (SDV) \cite{sdv1}. 

To test autonomous driving software, conventional evaluation methods include Software-in-the-Loop (SiL) and Hardware-in-the-Loop (HiL). Compared to real-world testing, these evaluation methods stand out for their highly reproducible environments, safe testing frameworks, cost-effectiveness, and accelerated development cycles. Nevertheless, traditional SiL and HiL setups often fall short in capturing the full dynamics of the central server-vehicle interactions, particularly when distributed ECUs are replaced by software-defined architectures. Therefore, it is preferred to test the vehicle as a whole component in the validation process. The Vehicle-in-the-Loop (ViL) method is used to validate the adaptability of the physical characteristics, including vehicle dynamics, from simulation to reality in early development phases. \cite{milestone}.  

The vehicle is tested under the "Hybrid Testing" methodology, in which the real vehicle is combined with a virtual one in a co-simulation framework to analyze driving performance in virtually created scenarios \cite{VEHIL_traffic}. The virtual vehicle in the simulation environment is, in this case, the digital twin of the real vehicle. The ViL methodology is applicable for the validation of both ADAS functions and fully autonomous driving vehicles.

A distributed E/E architecture results in repeated flashing of new software to the corresponding ECUs when autonomous driving functions need to be changed, which greatly increases development effort. In our work, it is not necessary to consider individual ECUs. Instead, the vehicle is tested as a whole using ViL and digital twin methodology. The key contributions of this paper include:

\begin{itemize}
\item A novel ViL validation setup with a test bench for the autonomous vehicles with a centralized E/E architecture.
\item The simulation environment for validation, the synchronized digital twin of the vehicle on the test bench, and the digital twin of the object.
\item Integration of autonomous driving algorithms into the vehicle validated on the test bench.
\item Validation tests with both manual-drive and autonomous driving functions, in which the real sensors are integrated.
\end{itemize}

\section{RELATED WORK}
In this section, relevant literature regarding the shift to centralized E/E architecture and the Vehicle-in-the-Loop (ViL) validation methods is introduced. %

\subsection{Centralized E/E structure}
Automotive E/E structures have experienced multiple stages of improvement over the past years. The intuitive point-to-point connections are established in the early stage due to the limited number of functional ECUs \cite{review}. This straightforward solution was later no longer applicable with the increasing number of ECUs. With the increasing demand for autonomous driving, functions in automotive software, as well as the corresponding ECUs, have expanded significantly. The bottlenecks include the demand for massive data processing and the need for data sharing and coordinated control between automotive systems \cite{requirement}. The increasing requirements result in the E/E architecture equipped with a centralized gateway connected to different subnetworks so that it converts various protocols and manages network traffic, which is already adopted by automobile manufacturers such as Volkswagen \cite{vw}, BMW \cite{bmw1}, and Audi \cite{audi}. Nevertheless, the huge amount of data that passes through leads to a high load on the gateway, causing bandwidth and latency problems \cite{gatewayload}. The Domain Control Units (DCU) were therefore proposed to address the problem caused by high gateway load, which provides isolation of the ECU based on similar functions and communication access \cite{requirement}. Additionally, to reduce cabling costs, Brunner \cite{zone} proposed a zone-based structure, where components are integrated based on their physical location within the vehicle.

Furthermore, with the development of the autonomous driving functionalities, which require High-Performance Computing, the centralized E/E architecture is proposed in the context of artificial intelligence, neural networks, cloud, and OTA updates \cite{modelling}. There have been several attempts to implement an individual centralized E/E structure. For example, BMW \cite{BMW_HPC} introduced the 4 "Superbrains" high-performance computers, which are responsible for the infotainment, automated driving, driving dynamics (via the Heart of Joy), and basic functions in the car respectively, and the hardware and software are decoupled to develop SDV. Other attempts for high-performance central computing include In Car Application Server (ICAS) from Volkswagen, SPA2 from Volvo, and FACE from Renault \cite{cybersecurity, visualizing}.

Apart from the OEMs from the industry, research and academia have also proposed several solutions for the centralized automotive architecture. Kirchner et al. \cite{autoframe} introduced the Autoframe structure, in which a scalable, modular, and safe automotive deployment framework with a hardware-abstraction layer is proposed for SDV. Another example is APIKS \cite{apiks}, a software platform based on ROS 2. It is designed for rapid prototyping and efficient validation of autonomous vehicle software within SDVs. However, there still lacks an approach that can test the centralized architecture in a realistic environment before the vehicle operates on the real road.

\subsection{Vehicle-in-the-Loop validation}
To simulate the dynamic conditions of the roads, two mainstream ViL test methods are used: the Field-based ViL and the Testbench-based ViL \cite{survey}. The vehicle operates directly on the physical roads in the Field-based ViL approach, while a laboratory test bench is utilized for the Testbench-based ViL. The Testbench-based ViL has the advantages of real sensor data input, lower space requirements, and reproducibility. Gietelink et al. \cite{VEHIL} used the "Vehicle-Hardware-in-the-Loop" (VEHIL) with a chassis dynamometer to perform functional testing and fault diagnosis for ADAS applications. The "DrivingCube" proposed by Schyr et al. \cite{avl} is another ViL testing framework for efficient validation of AV applications using a powertrain dynamometer. A steerable VEhicle-in-the-Loop (VEL) test bench validation methodology was proposed for the reliability of automated driving functions \cite{kit, radar}. Lee and Won \cite{fullscale} also proposed a full-scale test bench-based ViL method for autonomous vehicles, in which the sensor data are transmitted through over-the-air (OTA) stimulation or sensor data injection.

Nevertheless, an integrated and straightforward framework for a fully autonomous driving software development process, from implementation to validation, for vehicles with centralized E/E architecture still lacks. The existing solutions focus on testing individual ECUs of the conventional E/E architecture, and changing the algorithm itself requires tedious flashing of each ECU. Therefore, we present a novel closed-loop development and testing framework that bridges the gap between implementation and validation for algorithm development on a central car server using ViL. Our proposed work is similar to that introduced in \cite{fullscale}, but the concentration is placed on camera data processing and the efficient "Implementation - Realistic Validation" (I-RV) framework based on a central car server.

\section{Methodology}

\begin{figure*}[htb]
    \vspace{1em}
    \centering
    \includegraphics[width=0.96\textwidth]{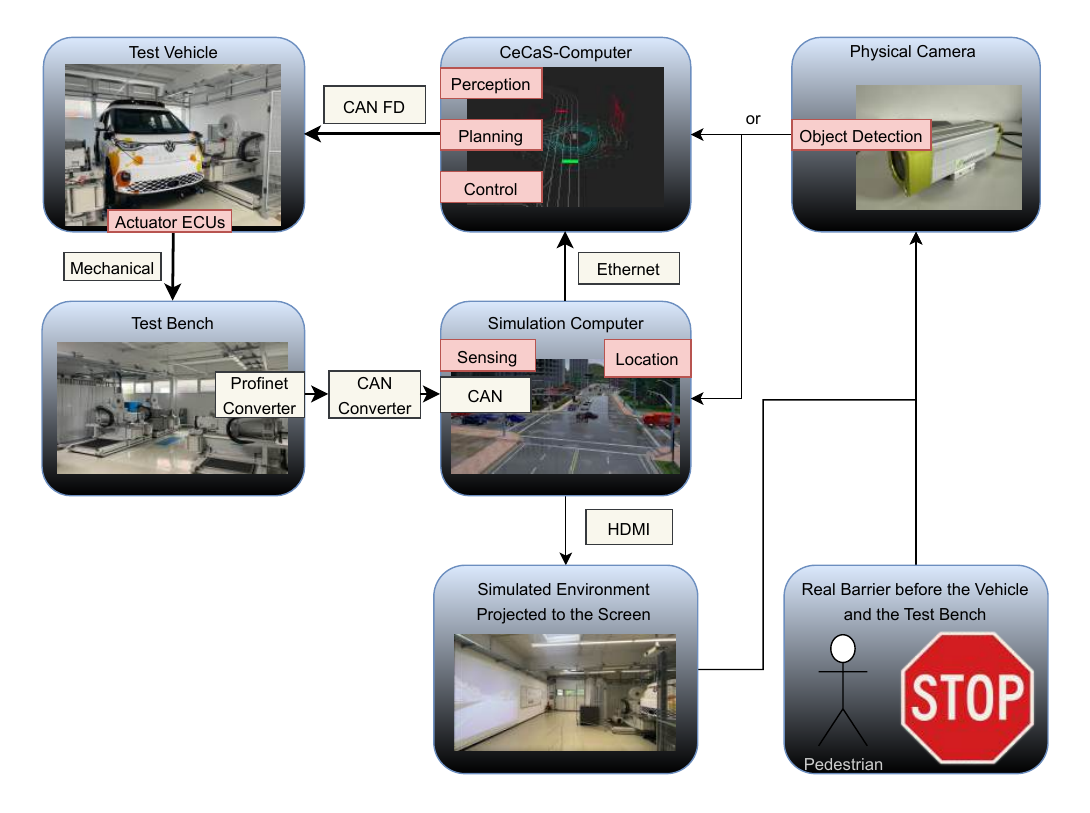}
    \caption{Overview of the proposed ViL validation framework with Central Car Server (CeCaS)-Computer as an alternative to the in-vehicle central car server. The virtual vehicle in the simulation environment serves as the digital twin of the test vehicle on the test bench.}
    \label{fig:setup}
\end{figure*}

\subsection{Hardware Architecture}
The significant hardware in our development framework includes: 
\begin{itemize}
\item A reconstructed rear-wheel-drive Volkswagen ID Buzz with external control sockets for driving algorithms.
\item A powertrain dynamometer vehicle test bench provided by SIEMENS for driving dynamics simulation of the test vehicle, supported by SIEMENS CATS-TC500 control system.
\item A Simulation Computer for CARLA \cite{carla}, in which the digital twin of the real test vehicle is spawned.
\item A High Performance Computer - the Central Car Server (CeCaS)-Computer, which serves as the central car server for autonomous driving functions.
\item A Basler Ace 2 Power over Ethernet (PoE) camera as the external physical sensor connected to the Simulation Computer and CeCaS-Computer.
\end{itemize}

\begin{figure}[htb]%
    \centering%
    %
    \includegraphics[width=40mm]{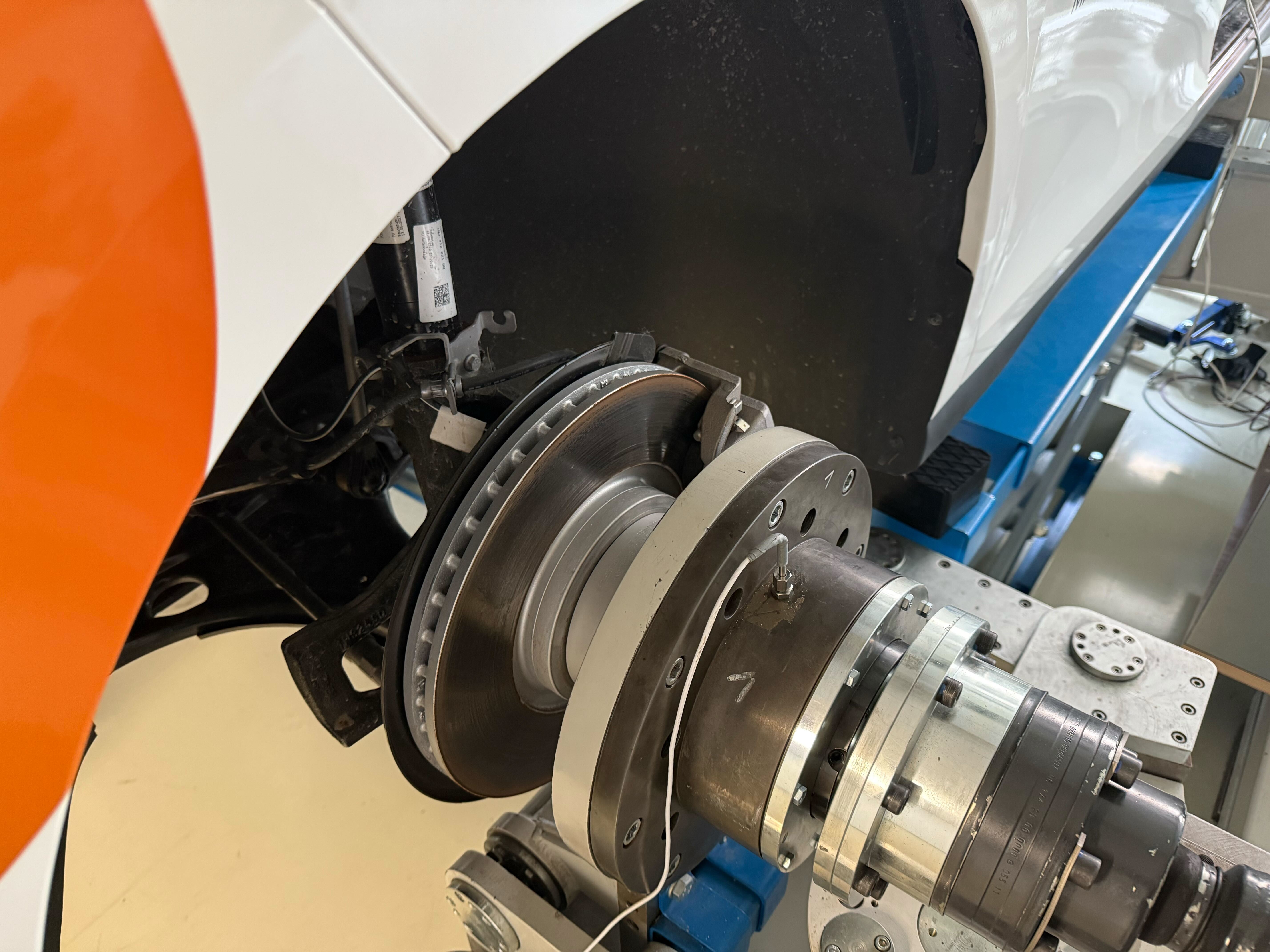}%
    \hspace*{1mm}%
    %
    \includegraphics[width=40mm]{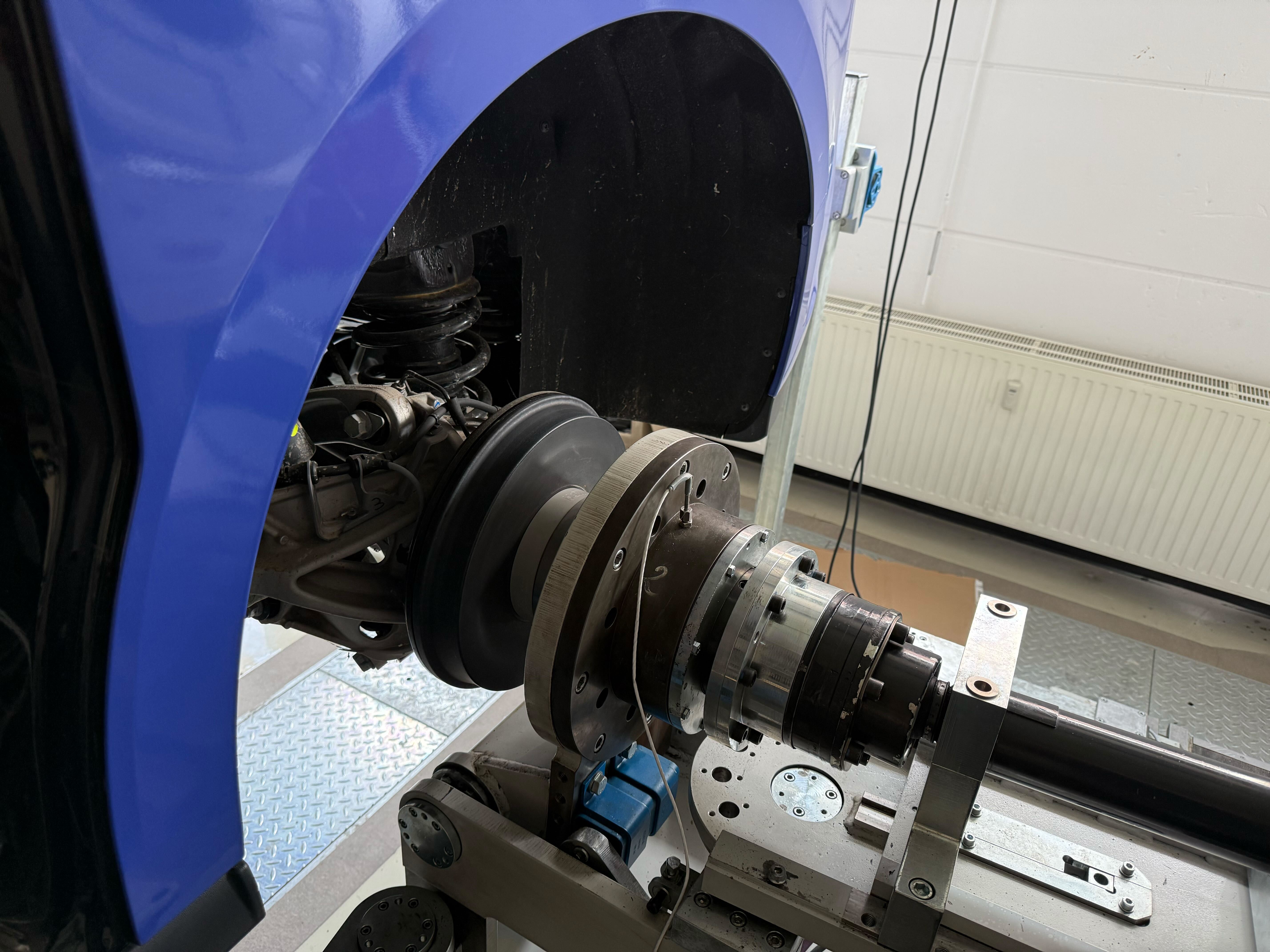}
    \caption{Mounting points of the front (left) and rear (right) axle to the test bench with adapters.}%
    \label{fig:adapters}%
\end{figure}

The overall structure of the hardware setup is visualized in \cref{fig:setup}. The test vehicle with unmounted wheels is mechanically connected to the test bench through the powertrain. The mounting points are illustrated in \cref{fig:adapters}. Consistent with conventional electric vehicle architectures, the front axle is equipped with disc brakes, whereas the rear axle employs drum brakes. Therefore, different adapters are necessary for adaptation to the test bench for different axles. The test bench provides the test vehicle with a realistic driving environment by generating rolling resistance torque in the longitudinal direction and self-aligning torque in the lateral direction. As the test vehicle is electric, exhaust gas emissions do not occur. Therefore, dedicated equipment for in-house exhaust gas accumulation was not necessary. A wall box is mounted next to the vehicle for an efficient and continuous testing process. In addition, the kinematic behavior of the vehicle is measured by the test bench and transmitted via Profinet-CAN to the simulation computer, where the digital twin of the test vehicle operates within a predefined scenario of the simulated environment. CARLA \cite{carla} was chosen because it is open-source, highly flexible, and easily integrates with the central car server and digital twin environment, enabling rapid prototyping and reproducible research. The hardware setup of the simulation computer and CeCaS-Computer is listed in \cref{tab:hardware}. Compared with other real-time control prototyping hardware, such as dSPACE or Speedgoat, our CeCaS-Computer is more suitable for centralized, high-performance, multi-sensor, and software-defined vehicle testing. Although the hardware setup natively supports multimodal sensors such as LiDAR and radar at the simulator and middleware levels, these sensors were not activated in the reported experiments in order to isolate the evaluation to perception-planning-control coupling and to ensure reproducibility and real-time determinism on the physical test bench. The test vehicle exhibits realistic powertrain dynamics, it remains nevertheless physically stationary on the test bench. The simulated environment is also projected onto the screen in front of the vehicle, which is visible to the driver sitting in the cockpit.

\begin{table}[h]
\centering
\footnotesize
\begin{tabular}{|l|l|l|}
\hline
\textbf{Name} & Simulation Computer & Central Car Computer \\ \hline
\textbf{CPU} & AMD TRP 5995WX & AMD TRP 5955WX \\ 
 & 64 Cores at 2.7 GHz & 16 Cores at 2.7 GHz \\ \hline
\textbf{GPU} & 6 x RTX 4090 24 GB VRAM & RTX 4090 24 GB VRAM \\ \hline
\textbf{RAM} & 256 GB DDR4-3200 RAM & 64 GB DDR4-3200 RAM \\ \hline
\end{tabular}
\caption{Hardware setups - The simulation computer is equipped with six graphics processors to handle the high computational load of simulating numerous sensors, whereas the vehicle’s central computer provides significantly less computing power in order to emulate a realistic operational scenario.}
\label{tab:hardware}
\end{table}



The autonomous driving functions implemented in the central car server rely on sensor setups, which can be either virtual or real. On the one hand, virtual sensors can be directly integrated into the simulation environment on the simulation computer, and the perceived information is then transmitted directly via Ethernet to the CeCaS-Computer. On the other hand, real sensor setups can also be included for the development of the autonomous driving algorithms. In this case, the Basler camera, for example, is installed for the demonstrated use case. The camera detects information either from the projected simulation environment or from real objects in front of the vehicle. The signal is then sent to the CeCaS-Computer for further processing. Alternatively, the signal indicating the detected objects can also be sent to the simulation computer, where the digital twins of the objects are generated in the simulation environment. The autonomous driving algorithms are implemented on the CeCaS-Computer for centralized control purposes. The test vehicle is controlled by the signal calculated by the CaCaS-Computer.

\subsubsection{Signal Frequency}
In the development process of autonomous driving algorithms, the CARLA simulator operates in synchronous mode. The CARLA world ticks once all the algorithmic calculations have completed. Therefore, it is assumed in the implementation that the hardware will not be the bottleneck in the future application, allowing the developer to concentrate on the algorithm itself.

The communication between CARLA simulation and object detection with the physical camera is via TCP. The physical camera sends the detection signal at a constant rate, and the CARLA world does not wait until the next signal. The camera detection rate is higher than CARLA's tick rate.

\subsubsection{Vehicle External Control}
For the control of the car via external software, it is necessary to modify the existing vehicle to enable the CeCaS system as a self-driving system (SDS) to communicate with the legacy ECUs and bus systems of the vehicle and to enable multiple operating modes.

To enable a prototypical SDS to communicate with the vehicle, an interface hardware called the vehicle motion gateway, consisting of a redundant set of two gateway ECUs, was implemented in the car. These gateway ECUs established communication with the vehicle legacy ECUs, which perform the actual vehicle control, such as acceleration, deceleration, and steering. Full hardware and software redundancies of the vehicle motion gateways are part of the functional safety concept to ensure a safe operation of the vehicle. Both gateways provide an individual private CAN interface towards the SDS, where one channel is considered to be the primary and default, while the other provides a fallback option to control the vehicle with limited functionality. To ensure safe and consistent data transmission, CAN signals of the vehicle control interface are end-to-end protected, and the values are inspected for validity. In addition, the vehicle's functional safety concept involves manual override and operational fallback states, as well as an emergency stop concept with the dynamometer test bench. The SDS can run on a dedicated PC or on a high-performance compute module as a hardware component of the central car server. With this control concept, a safe automated operation of the vehicle in the powertrain dynamometer test bench is developed.

\subsection{Software Solutions}
Our proposed software development process undergoes three steps:
\begin{enumerate}
\item \textbf{Internal test}, with which the developed software is only tested on the same hardware platform, with the environment simulated. It corresponds to the SiL method in the conventional process.
\item \textbf{External test}, with which the developed software is implemented on an isolated hardware, with the environment simulated on another hardware. Ethernet is used as the communication protocol. It corresponds to the HiL method in the conventional process.
Misalignment compared to the internal test stage may occur due to timing behavior, communication effects, and platform-dependent execution differences, and these can be mitigated accordingly before the software is deployed in the ViL stage.
\item \textbf{ViL test}, with which the software is tested on the central car server on the test vehicle on the test bench, and the scenario is simulated with the High Performance Computer.
Additional misalignment sources compared to external test arise due to in-vehicle network delays, middleware scheduling effects, actuator dynamics, and discrepancies between simulated and real vehicle behavior, as well as accumulated latency in the closed-loop system. The ViL stage exposes these effects under realistic execution conditions, enabling final timing analysis, calibration, and interface refinement before deployment.
\end{enumerate}

To realize the digital twin in the CARLA simulation environment and establish a smooth data-processing pipeline, several challenges exist.

\subsubsection{Digital Twin in the Simulation Environment}
The digital twin test method enables the combined testing of virtual and real environments without physical risks. The digital vehicle in the simulation environment behaves identically to its physical twin on the test bench. Scenarios can therefore be generated in the simulation environment without changing the physical test settings. A "hybrid" sensor data source is possible. The sensor data to be processed in this case comes from the simulated sensors. As an alternative, the real sensors can also be applied in the scenario, with the physical camera as an example, and the CeCaS-Computer processes the sensor signal from the real camera. 

There is always a trade-off between the accurate vehicle dynamics and high-fidelity sensor simulations \cite{racing}. The problem of the sim-to-real gap is always a concern for researchers. Therefore, the CARLA vehicle model in our setup is replaced by the test bench. The test is conducted with a realistic hardware framework. According to the specific vehicle's kinematic and dynamic parameters, the test bench is configured to generate realistic friction forces in both longitudinal and lateral directions.

The CeCaS-Computer sends the calculated control signal, such as throttle position, braking, steering angle, and the turn signal, to the test vehicle, and the vehicle follows these signals as long as the external control mode is activated. The vehicle can also be controlled manually by a human driver with normal driving maneuvers.

\subsubsection{Sensor Signal Transmitted from Simulation Computer to CeCaS-Computer}
The central computing software runs on the CeCaS-Computer, and the general pre-processed signal is sent from the simulation computer to it. Integrated simulated sensors include the RGB camera, the depth camera, the Lidar, and the radar. The RGB camera provides the RGB image directly, while the depth camera tells the distance of each pixel from the camera. The 3D point cloud is generated by simulating a rotating Lidar implemented using ray casting. The simulated radar sensor creates a 2D point map of the elements in sight.

\section{Experiments}
To validate the effectiveness and robustness, both manual-drive, in which a human driver controls the vehicle behavior using the steering wheel and pedals, and autonomous driving scenarios are included in the experiments. The algorithms are developed and implemented in the central CeCaS-Computer, and the output of the algorithm is sent to the test vehicle for external control.

\subsection{Manual-Drive}
\begin{figure}[!htbp]
    \centering%
	\includegraphics[width=70mm]{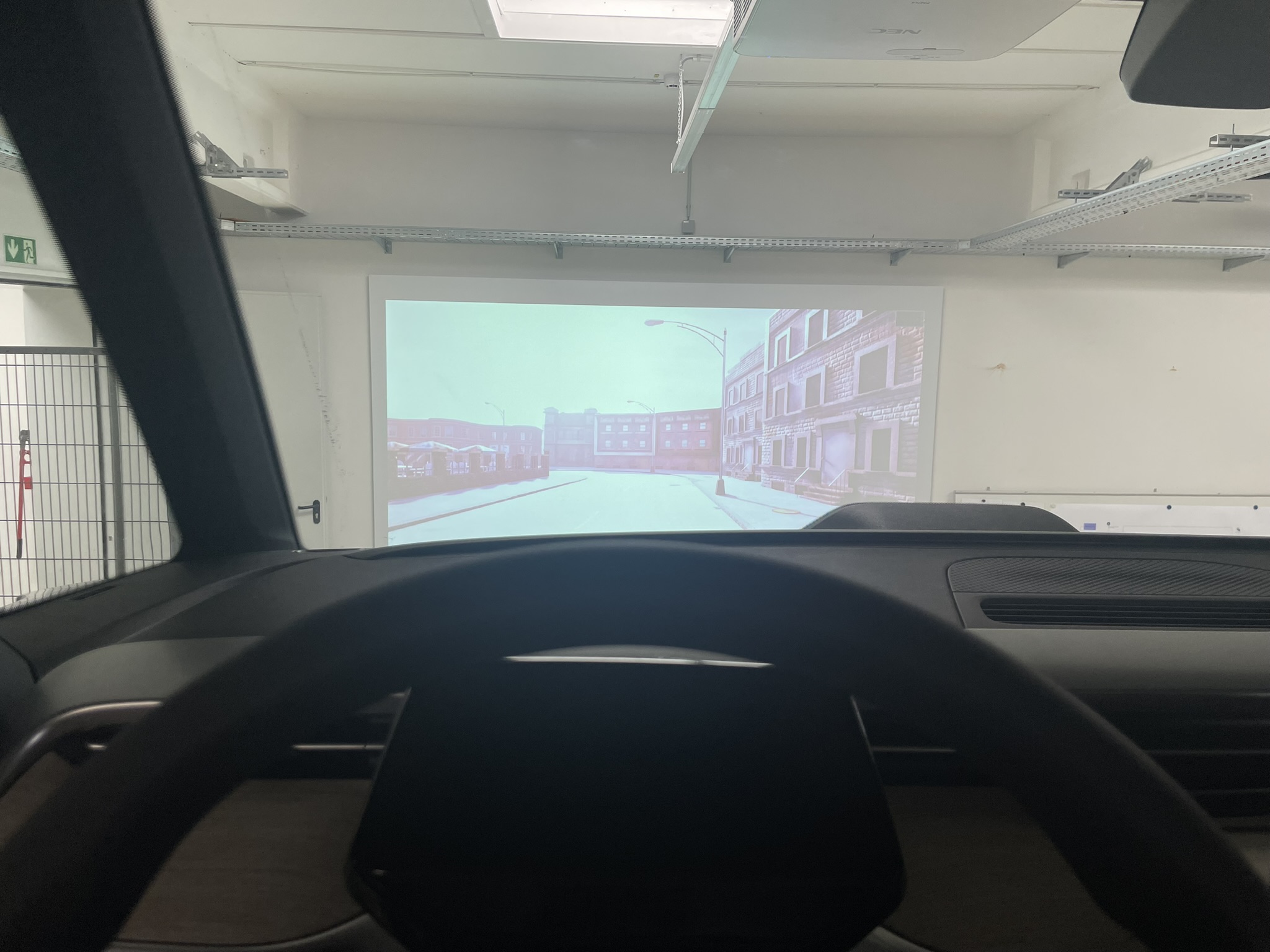}
	\caption{View from the cockpit of the test vehicle under manual-drive scenarios.}
	\label{fig:cockpit}
\end{figure}

The manual driving functionality of the test vehicle was evaluated using a high-fidelity simulation framework based on CARLA’s standard urban and highway maps. The test setup operates similarly to a driving simulator, allowing a driver to control the physical test vehicle while the vehicle’s behavior is simultaneously mirrored in a digital twin within the simulation environment. Other traffic participants, including vehicles and pedestrians, can be spawned in the simulation environment to enable realistic interactions and assess the ego vehicle’s behavior under dynamic traffic conditions. All vehicle maneuvers, including steering, acceleration, and braking, executed on the physical test bench were continuously projected onto the corresponding digital twin in the CARLA world, ensuring precise synchronization between the real-world and simulated environments. This approach enables a detailed analysis of vehicle dynamics and driver interactions under controlled yet realistic conditions, while leveraging the flexibility of a virtual environment to safely reproduce complex traffic scenarios. The driver's view from the cockpit is shown in \cref{fig:cockpit}.

\subsection{Autonomous Driving}
In this driving scenario, the control signals are provided by the central CeCaS-Computer, and the test vehicle is controlled externally. Two use cases are utilized for validation and demonstration:

\begin{itemize}
\item The Adaptive Cruise Control (ACC) with Lane Keeping Assist (LKA).
\item The automatic Emergency Brake function with a physical camera mounted on the test vehicle.
\end{itemize}

The ACC and LKA are treated as the basic autonomous driving functions, and the Emergency Brake is implemented based on these two fundamental functions.

\subsubsection{ACC and LKA}
The autonomous driving functions are implemented separately in the lateral and longitudinal directions. The applied algorithm consists of three modules: perception, planning, and control. YOLO\cite{yolo} is applied for the detection of the drivable area and the middle curve with an RGB and depth camera. The middle curve is subsequently converted into a list of CARLA waypoints, which the vehicle should follow. The corresponding steering angle should then be calculated to let the ego-vehicle follow the generated ideal curve. Various control methods are applied, including the conventional geometric method, Model Predictive Control, and Reinforcement Learning \cite{chengdong}.

In the longitudinal direction, the distance between the ego vehicle and the preceding vehicle is estimated using YOLO \cite{yolo}. Based on this distance, the desired acceleration is computed, from which the corresponding throttle command is subsequently derived.

\begin{figure}[!htbp]
    \centering%
	\includegraphics[width=70mm]{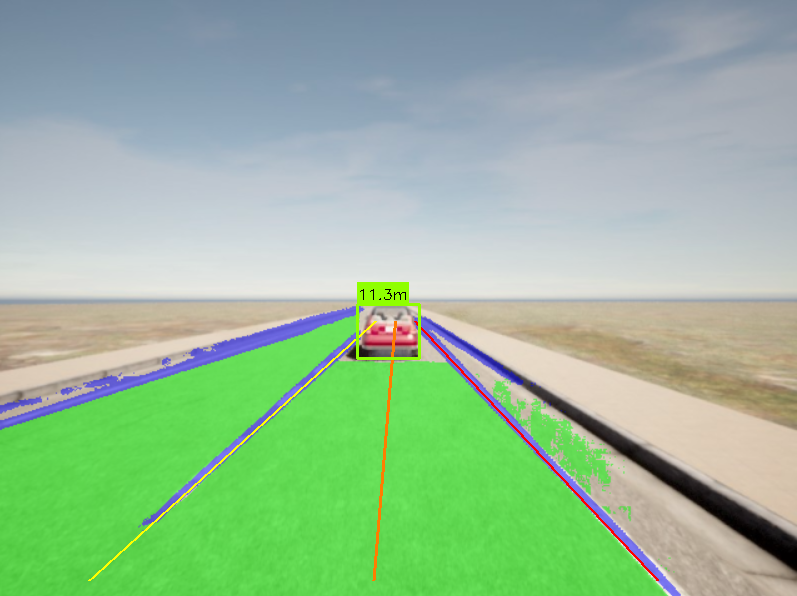}
	\caption{Vehicle front camera in the ACC scenario.}
	\label{fig:acc}
\end{figure}

In \cref{fig:acc}, the vehicle-following scenario is illustrated, and the corresponding drivable area is painted in green.

\subsubsection{Emergency Brake}
\begin{figure}[htb]%
    \centering%
    %
    \includegraphics[width=40mm]{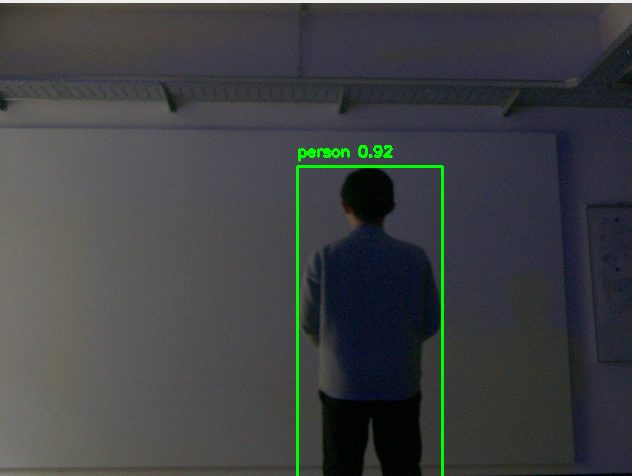}%
    \hspace*{1mm}%
    %
    \includegraphics[width=40mm]{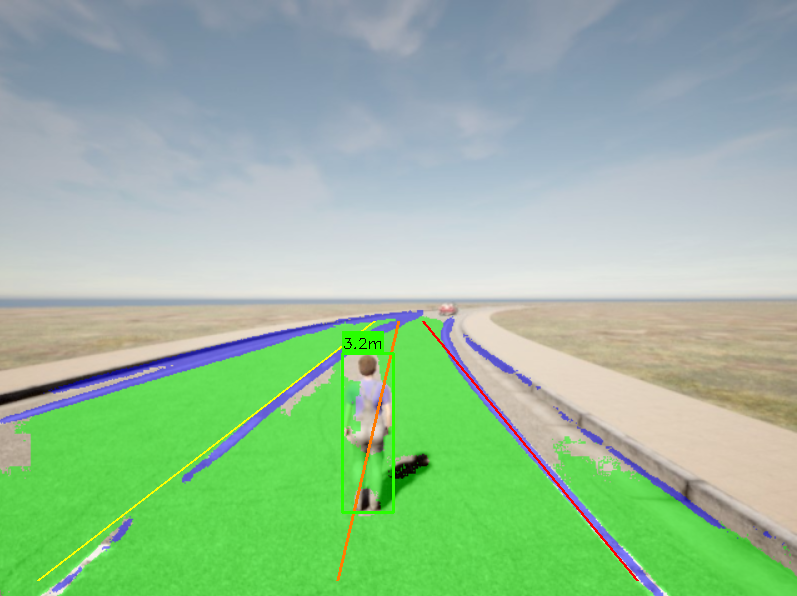}
    \caption{Left: Front camera view on the test vehicle with the person detected.
    Right: Virtual simulation scenario and digital twin of the person spawned on the track when the object is detected with the real camera.}%
    \label{fig:personandeb}%
\end{figure}

An external sensor is employed to provide realistic perception signals to the test setup. In this configuration, a front-facing camera is mounted on the vehicle to enable object detection. The sensor data are processed using a YOLO-based detection algorithm \cite{yolo}. During the test case, a digital twin of the detected person is spawned on the simulated track in front of the ego vehicle, as illustrated in \cref{fig:personandeb}. Using ACC as the baseline scenario, both the simulated and physical vehicles perform an emergency braking maneuver upon object detection. During each synchronous simulation step, the controller uses the most recent perception result whose timestamp is closest to the current simulation time. Detection results are held constant between updates and evaluated at every control cycle. This scenario serves not only as a test case for system validation, but also as a safety mechanism to mitigate potential risks when unforeseen objects approach the test bench.




\section{Analysis}
In this section, the test results of the test bench setup are divided into two parts: the structural results and the performance results.

\subsection{Structural analysis}
Data should be transmitted between different modules of the hardware setups. In the simulation environment, CARLA operates in synchronous mode to synchronize the virtual vehicle with the real system. The control signals for vehicle acceleration and steering angles are transmitted at a rate of 20 milliseconds for continuous and smooth control. Other signals with less real-time demands, such as turn signals, are transmitted in a 50-millisecond interval.

The camera detection algorithm runs with its own frequency. Five frames are captured per second, and they are not synchronized with the simulation environment, which resembles the real case when the camera is mounted on a real vehicle. In each detection loop, the camera sends a signal with the detected information to the hardware setup.

\subsection{Performance analysis with the demonstrating algorithm}
\begin{figure}[!htbp]
    \centering%
	\includegraphics[width=70mm]{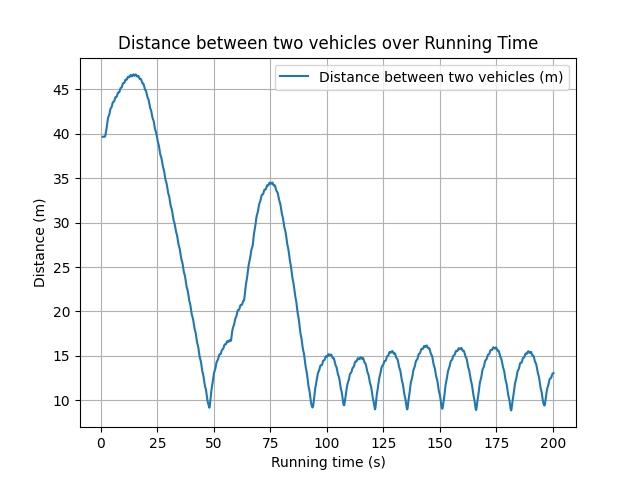}
	\caption{Distance between two vehicles when the ego-vehicle approaches.}
	\label{fig:accb}
\end{figure}

The ACC and LKA are implemented for demonstration purposes. \cref{fig:accb} depicts the performance of the car-following ACC algorithms on our test setup. The ego-vehicle starts with the initial velocity of 0 km/h and begins to approach the leading vehicle. The average lateral error, which is defined as the shortest distance between the virtual vehicle's position and the center line from the map's ground truth in CARLA as waypoints, remains under 0.05 m. Quantitative results demonstrate the functionality of the ViL test setup and the autonomous driving algorithm.

\begin{figure}[htb]%
    \centering%
    %
    \includegraphics[width=45mm]{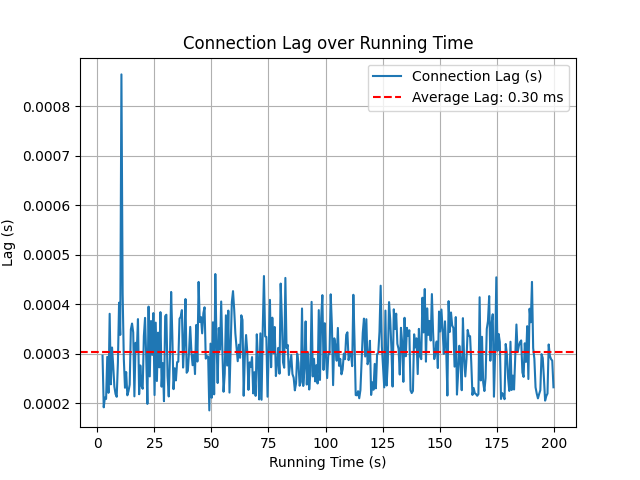}%
    \hspace*{1mm}%
    %
    \includegraphics[width=45mm]{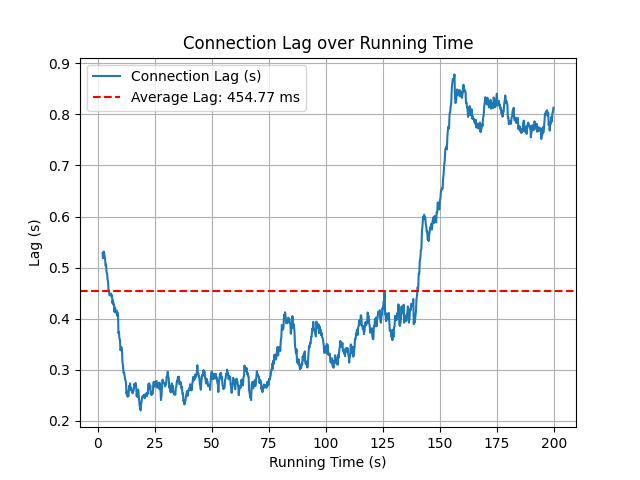}
    \caption{Connection latency with FPS of the camera = 2 (left) and 5 (right)}%
    \label{fig:latency}%
\end{figure}

Repeated tests are conducted in the emergency braking scenario. The latencies of the emergency brake system trigger are measured during operation after the person is detected by the camera in front of the vehicle. Latencies with different camera capture frequencies are demonstrated in \cref{fig:latency}. The connection lags are highly dependent on the FPS of the camera detection due to the computational resources required for object detection. Because the car-following algorithm operates during the second part of the simulation, the latency increases to a higher value due to the growing consumption of computational resources.



\section{Conclusions}
We propose a ViL framework with both virtual and real sensors for the validation and testing of new E/E architecture approaches based on a central car server (CeCaS). The essential parts in the setup include the externally controllable test vehicle, a vehicle test bench for dynamic and kinematic measurement of test vehicle status, a powerful simulation computer for environment simulation and virtual sensor signal generation, a CeCaS-Computer used for the central car server, and the real sensors mounted on the test vehicle. The closed-loop digital twin test methodology has been built and implemented, combining the advantages of both virtual testing in the simulation environment and the dynamic performance of the real vehicle. The vehicle can either be driven manually with a human included, or fully autonomously with the external control sockets. With the real sensors on the vehicle, not only can the vehicle itself be mirrored into the virtual simulation environment, but objects such as people in front of the vehicle also have their digital twins. The reproducible and cost-effective test setup enables the automotive industry not only to evaluate newly developed functions and technologies on existing vehicle model platforms, such as modern centralized E/E architectures, but also to validate innovations on next-generation vehicles with redesigned platforms in the early development process.

The proposed vehicle-in-the-loop test bench provides a seamless “code-to-road” workflow, offering a highly realistic testing environment for autonomous driving algorithms. This setup significantly streamlines the development process, as algorithms can be directly deployed and evaluated on the central vehicle server without the need to individually flash, configure, and test separate ECUs. By eliminating these intermediate steps, the platform not only reduces development effort and complexity but also accelerates iterative testing and validation, enabling rapid refinement of autonomous functionalities in conditions closely replicating real-world scenarios. This approach represents a pivotal step toward more efficient, integrated, and scalable autonomous vehicle development.

\section{Future Work}
In future research work, there are areas that can be refined and improved. First, a broader range of autonomous driving algorithms and additional, more complicated scenarios can be implemented to enable more comprehensive testing and validation across diverse operational scenarios. By connecting the CeCaS-Computer with a centralized vehicle IT infrastructure, such as a central ECU, ZoneECUs, and Ethernet Backbone, the runtime performance of the entire system can be evaluated. In addition, the test bench can be expanded with additional virtual and real sensing modalities, such as LiDAR and radar, including the corresponding sensor data generation and synchronization pipelines. This enhancement will allow the platform to more accurately replicate the heterogeneous perception environments of modern automated vehicles. Furthermore, novel sensor concepts and hardware can also be tested safely on the test bench, utilizing both the vehicle's hardware and the autonomous driving software. As the manual-drive function is available, human beings can also be involved in the ergonomic study within the frame of Human-in-the-Loop. The ultimate step would involve transferring the developed algorithms to real-world driving tests and comparing the on-road data with that obtained in the ViL environment. This comparison will enable the quantitative assessment of the platform's fidelity, robustness, and transferability to actual driving conditions.

\begingroup
\setstretch{0.85} 
{\small
\printbibliography}

@proceedings{.2007,
 year = {2007},
 title = {2007 IEEE Intelligent Vehicles Symposium}
}

@proceedings{.2017,
 year = {2017},
 title = {2017 13th Workshop on Intelligent Solutions in Embedded Systems (WISES)}
}

@proceedings{.2020,
 year = {2020},
 title = {2020 IEEE Intelligent Vehicles Symposium (IV)}
}

@proceedings{.2024,
 year = {2024},
 title = {2024 IEEE 27th International Conference on Intelligent Transportation Systems (ITSC)}
}

@proceedings{.2024b,
 year = {2024},
 title = {2024 IEEE International Automated Vehicle Validation Conference (IAVVC)}
}

@proceedings{.2025,
 year = {2025},
 title = {2025 IEEE Intelligent Vehicles Symposium (IV)}
}

@proceedings{.2025b,
 year = {2025},
 title = {2025 IEEE Intelligent Vehicles Symposium (IV)}
}

@misc{apiks,
 author = {{Jo{\~a}o-Vitor Zacchi} and {Edoardo Clementi} and {N{\'u}ria Mata}},
 year = {2025},
 title = {APIKS: A Modular ROS2 Framework for Rapid Prototyping and Validation of Automated Driving Systems}
}

@article{audi,
 author = {{U{\u{g}}ur Keskin}},
 year = {2009},
 title = {In-vehicle communication networks:a literature survey},
 url = {https://api.semanticscholar.org/CorpusID:17671290},
 journal = {Virtual Reality}
}

@inproceedings{autoframe,
 crossref = {.2024},
 author = {Kirchner, Sven and Purschke, Nils and Wu, Chengdong and Khan, Muhammed Aqib and Dixit, Divye and Knoll, Alois C.},
 title = {AUTOFRAME - A Software-Driven Integration Framework for Automotive Systems},
 pages = {521--528},
 booktitle = {2024 IEEE 27th International Conference on Intelligent Transportation Systems (ITSC)},
 year = {2024},
}

@INPROCEEDINGS{avl,
  author={Schyr, Christian and Inoue, Hideo and Nakaoka, Yuji},
  booktitle={2022 International Conference on Connected Vehicle and Expo (ICCVE)}, 
  title={Vehicle-in-the-Loop Testing - a Comparative Study for Efficient Validation of ADAS/AD Functions}, 
  year={2022},
  volume={},
  number={},
  pages={1-8},
}

@proceedings{BargendeMichaelandReussHansChristianandWagnerAndreas.2021,
 year = {2021},
 title = {21. Internationales Stuttgarter Symposium},
 publisher = {{Springer Fachmedien Wiesbaden}},
 isbn = {978-3-658-33521-2},
 editor = {{Bargende, Michael and Reuss, Hans-Christian and Wagner, Andreas}}
}

@online{BMW_HPC,
 author = {{BMW Group}},
 year = {2025},
 title = {Four Superbrains for the Neue Klasse by BMW},
 url = {https://www.bmwgroup.com/en/news/general/2025/superbrains.html}
}

@article{bmw1,
 author = {Tuohy, Shane and Glavin, Martin and Hughes, Ciar{\'a}n and Jones, Edward and Trivedi, Mohan and Kilmartin, Liam},
 year = {2015},
 title = {Intra-Vehicle Networks: A Review},
 pages = {534--545},
 volume = {16},
 number = {2},
 journal = {IEEE Transactions on Intelligent Transportation Systems},
}

@inproceedings{carla,
  title = {{CARLA}: {An} Open Urban Driving Simulator},
  author = {Alexey Dosovitskiy and German Ros and Felipe Codevilla and Antonio Lopez and Vladlen Koltun},
  booktitle = {Proceedings of the 1st Annual Conference on Robot Learning},
  pages = {1--16},
  year = {2017}
}

@article{case,
 author = {Bandur, Victor and Selim, Gehan and Pantelic, Vera and Lawford, Mark},
 year = {2021},
 title = {Making the Case for Centralized Automotive E/E Architectures},
 pages = {1230--1245},
 volume = {70},
 number = {2},
 journal = {IEEE Transactions on Vehicular Technology},
}

@inproceedings{chengdong,
 crossref = {.2025},
 author = {Wu, Chengdong and Kirchner, Sven and Purschke, Nils and Knoll, Alois C.},
 title = {Autonomous Vehicle Lateral Control Using Deep Reinforcement Learning with MPC-PID Demonstration},
 pages = {2317--2324},
 booktitle = {2025 IEEE Intelligent Vehicles Symposium (IV)},
 year = {2025},
}

@article{cybersecurity,
 author = {Tany, Nadera Sultana and Suresh, Sunish and Sinha, Durgesh Nandan and Shinde, Chinmay and Stolojescu-Crisan, Cristina and Khondoker, Rahamatullah},
 year = {2022},
 title = {Cybersecurity Comparison of Brain-Based Automotive Electrical and Electronic Architectures},
 volume = {13},
 number = {11},
 issn = {2078-2489},
 journal = {Information},
}

@article{daimler,
title = {Centralization potential of automotive E/E architectures},
journal = {Journal of Systems and Software},
volume = {219},
pages = {112220},
year = {2025},
issn = {0164-1212},
author = {Lucas Mauser and Stefan Wagner}
}

@inproceedings{fullscale,
 crossref = {.2024b},
 author = {Lee, Yong-Ha and Won, Jong-Hoon},
 title = {Study on Full Scale Testbench Based Vehicle-In-The-Loop Simulation Testing for Automated Vehicles},
 pages = {1--6},
 booktitle = {2024 IEEE International Automated Vehicle Validation Conference (IAVVC)},
 year = {2024},
}

@article{gatewayload,
 author = {{Lo Bello}, Lucia and Patti, Gaetano and Leonardi, Luca},
 year = {2023},
 title = {A Perspective on Ethernet in Automotive Communications---Current Status and Future Trends},
 volume = {13},
 number = {3},
 issn = {2076-3417},
 journal = {Applied Sciences},
}

@inproceedings{kit,
 crossref = {BargendeMichaelandReussHansChristianandWagnerAndreas.2021},
 author = {{Han, Chenlei and Seiffer, Alexander and Orf, Stefan and Hantschel, Frank and Li, Shiqing}},
 title = {Validating Reliability of Automated Driving Functions on a Steerable VEhicle-in-the-Loop (VEL) Test Bench},
 pages = {546--559},
 publisher = {{Springer Fachmedien Wiesbaden}},
 isbn = {978-3-658-33521-2},
 editor = {{Bargende, Michael and Reuss, Hans-Christian and Wagner, Andreas}},
 booktitle = {21. Internationales Stuttgarter Symposium},
 year = {2021},
 address = {Wiesbaden}
}

@inproceedings{milestone,
 crossref = {.2007},
 author = {Bokc, Thomas and Maurer, Markus and Farber, Georg},
 title = {Validation of the Vehicle in the Loop (VIL); A milestone for the simulation of driver assistance systems},
 pages = {612--617},
 booktitle = {2007 IEEE Intelligent Vehicles Symposium},
 year = {2007},
}

@article{modelling,
 author = {{Alessio Bucaioni} and {Patrizio Pelliccione} and {Saad Mubeen}},
 year = {2024},
 title = {Modelling centralised automotive E/E software architectures},
 pages = {102289},
 volume = {59},
 issn = {1474-0346},
 journal = {Advanced Engineering Informatics},
}

@inproceedings{racing,
 crossref = {.2025b},
 author = {Brunner, Maurice and Ghignone, Edoardo and Baumann, Nicolas and Magno, Michele},
 title = {R-CARLA: High-Fidelity Sensor Simulations with Interchangeable Dynamics for Autonomous Racing},
 pages = {2558--2564},
 booktitle = {2025 IEEE Intelligent Vehicles Symposium (IV)},
 year = {2025},
}

@article{radar,
 author = {Diewald, Axel and Kurz, Clemens and Kannan, Prasanna Venkatesan and Gie{\ss}ler, Martin and Pauli, Mario and G{\"o}ttel, Benjamin and Kayser, Thorsten and Gauterin, Frank and Zwick, Thomas},
 year = {2021},
 title = {Radar Target Simulation for Vehicle-in-the-Loop Testing},
 pages = {257--271},
 volume = {3},
 number = {2},
 issn = {2624-8921},
 journal = {Vehicles},
}

@article{requirement,
 author = {Zhu, Hailong and Zhou, Wei and Li, Zhiheng and Li, Li and Huang, Tao},
 year = {2021},
 title = {Requirements-Driven Automotive Electrical/Electronic Architecture: A Survey and Prospective Trends},
 pages = {100096--100112},
 volume = {9},
 journal = {IEEE Access},
}

@article{review,
 author = {Wang, Wenwei and Guo, Kaidi and Cao, Wanke and Zhu, Hailong and Nan, Jinrui and Yu, Lei},
 year = {2024},
 title = {Review of Electrical and Electronic Architectures for Autonomous Vehicles: Topologies, Networking and Simulators},
 pages = {82--101},
 volume = {7},
 number = {1},
 issn = {2522-8765},
 journal = {Automotive Innovation},
}

@online{sdv1,
 author = {{Dijaz Maric}},
 title = {When Software Takes the Wheel: The Future of Automotive Architecture},
 url = {https://lorit-consultancy.com/en/2024/10/when-software-takes-the-wheel-the-future-of-automotive-architecture/}
}

@article{state,
 author = {Guo, Zixuan and Koufos, Konstantinos and Dianati, Mehrdad and Woodman, Roger},
 year = {2025},
 title = {State-of-the-art virtualisation technologies for the centralised automotive E/E architecture},
 volume = {6},
 journal = {Frontiers in Future Transportation},
}

@article{survey,
 author = {Cheng, Jingjun and Wang, Zhen and Zhao, Xiangmo and Xu, Zhigang and Ding, Ming and Takeda, Kazuya},
 year = {2024},
 title = {A Survey on Testbench-Based Vehicle-in-the-Loop Simulation Testing for Autonomous Vehicles: Architecture, Principle, and Equipment},
 pages = {2300778},
 volume = {6},
 number = {6},
 journal = {Advanced Intelligent Systems},
}

@article{VEHIL,
 author = {{O. Gietelink} and {J. Ploeg} and {B. De Schutter} and {M. Verhaegen}},
 year = {2004},
 title = {VEHIL: A Test Facility for Validation of Fault Management Systems for Advanced Driver Assistance Systems},
 pages = {397--402},
 volume = {37},
 number = {22},
 issn = {1474-6670},
 journal = {IFAC Proceedings Volumes},
}

@inproceedings{VEHIL_traffic,
 crossref = {.2020},
 author = {Solmaz, Selim and Rudigier, Martin and Mischinger, Marlies},
 title = {A Vehicle-in-the-Loop Methodology for Evaluating Automated Driving Functions in Virtual Traffic},
 pages = {1465--1471},
 booktitle = {2020 IEEE Intelligent Vehicles Symposium (IV)},
 year = {2020},
}

@article{visualizing,
 author = {Phadnis, Tejas Pravin and Feyerabend, Nils and Axmann, Joachim},
 year = {2024},
 title = {Visualizing and analysing data-driven shift from decentralized to centralized automotive E/E architectures},
 pages = {453--462},
 volume = {4},
 journal = {Proceedings of the Design Society},
}

@article{vw,
 author = {Zeng, Weiying and Khalid, Mohammed A. S. and Chowdhury, Sazzadur},
 year = {2016},
 title = {In-Vehicle Networks Outlook: Achievements and Challenges},
 pages = {1552--1571},
 volume = {18},
 number = {3},
 journal = {IEEE Communications Surveys {\&} Tutorials},
}

@article{yolo,
 author = {{Joseph Redmon} and {Santosh Kumar Divvala} and {Ross B. Girshick} and {Ali Farhadi}},
 year = {2015},
 title = {You Only Look Once: Unified, Real-Time Object Detection},
 volume = {abs/1506.02640},
 journal = {CoRR}
}

@inproceedings{zone,
 crossref = {.2017},
 author = {Brunner, Stefan and Roder, Jurgen and Kucera, Markus and Waas, Thomas},
 title = {Automotive E/E-architecture enhancements by usage of ethernet TSN},
 pages = {9--13},
 booktitle = {2017 13th Workshop on Intelligent Solutions in Embedded Systems (WISES)},
 year = {2017},
}
\endgroup


\end{document}